\documentclass[a4paper, 10pt, conference]{ieeeconf}      
\usepackage{cite}
\usepackage{amsmath,amssymb,amsfonts}
\usepackage{algorithmic}
\usepackage{graphicx}
\usepackage{textcomp}
\usepackage{xcolor}
\usepackage{booktabs}

\IEEEoverridecommandlockouts                              
\title{\LARGE \bf
SelfieAvatar: Real-time Head Avatar reenactntment from a Selfie Video}

\usepackage{fancyhdr}
\begin{document}

\author{\parbox{16cm}{\centering
    {\large Wei Liang$^{1,2}$, Hui Yu$^2$, Derui Ding$^1$, Rachael E. Jack$^2$ and Philippe G. Schyns$^2$ }\\
    {\normalsize
    $^1$ Department of Control Science and Engineering, University of Shanghai for Science and Technology, Yangpu, Shanghai 200093, China\\
    $^2$  School of Psychology and Neuroscience, University of Glasgow, Hillhead Street, Glasgow G12 8QB, UK}}
    \thanks{This work was supported by the Horizon Europe Programme under the MSCA grant for ACMod (Grant No. 101130271, EP/Y03726X/1), European Research Council under the European Union’s Horizon 2020 research and innovation program [759796].}
}\pagestyle{plain}
\maketitle\thispagestyle{fancy}

\begin{abstract}

Head avatar reenactment focuses on creating animatable personal avatars from monocular videos, serving as a foundational element for applications like social signal understanding, gaming, human-machine interaction, and computer vision. Recent advances in 3D Morphable Model (3DMM)-based facial reconstruction methods have achieved remarkable high-fidelity face estimation. However, on the one hand, they struggle to capture the entire head, including non-facial regions and background details in real time, which is an essential aspect for producing realistic, high-fidelity head avatars. On the other hand, recent approaches leveraging generative adversarial networks (GANs) for head avatar generation from videos can achieve high-quality reenactments but encounter limitations in reproducing fine-grained head details, such as wrinkles and hair textures. In addition, existing methods generally rely on a large amount of training data, and rarely focus on using only a simple selfie video to achieve avatar reenactment. To address these challenges, this study introduces a method for detailed head avatar reenactment using a selfie video. The approach combines 3DMMs with a StyleGAN-based generator. A detailed reconstruction model is proposed, incorporating mixed loss functions for foreground reconstruction and avatar image generation during adversarial training to recover high-frequency details. Qualitative and quantitative evaluations on self-reenactment and cross-reenactment tasks demonstrate that the proposed method achieves superior head avatar reconstruction with rich and intricate textures compared to existing approaches.

\end{abstract}

\section{INTRODUCTION}

Reenacting human personal avatars from monocular videos has been an increasingly prominent field of study in computer vision\cite{wang2022morf}, human-machine interaction \cite{cao2023high} and psychological analysis \cite{Jack2012facial}, finding wide applications across various disciplines fields such as psychology and cognitive science\cite{Yang_Zhou_Wei_2024, Yan2024.05.06}, Virtual Reality (VR), Augmented Reality (AR)\cite{Zhao_Li_Xu_2022}, and other applications \cite{Yu2024sora, Ennis2013game, chen2022sofgan}. This task aims to reconstruct a faithful 3D avatar model with accurate shape and detailed appearance, providing 3D consistency, strong identity preservation, and controllable pose and expression. Learning detailed facial textures and generating realistic, high-fidelity avatars is a challenging task. For instance, the skin exhibits complex diffuse reflections, the eyes have highly reflective specular surfaces, and the hair features intricate geometry and fine details. Furthermore, the surrounding non-facial area near the head contributes significantly to enhancing the realism of the generated re-animation avatar video.

To reconstruct animatable personal avatars from a video, an early approach employed neural radiance fields (NeRF) \cite{Gafni2021NeRF} to generate personalized avatars with high-fidelity image rendering. More recent methods combined traditional 3D Morphable Models with supplementary networks to achieve more comprehensive expression and pose modeling for personalized avatars \cite{Chai2023hiface}. From a different perspective, some methods leveraged the powerful generation capability of generative adversarial networks (GAN) network to learn to create high-resolution and photo-realistic head avatars by incorporating latent space with 3DMM parameters. However, these methods often lack the ability to reconstruct fine details such as wrinkles and tend to produce smoother results due to their reliance on pre-trained GAN networks.

Tracking faces from videos and learning the 3D facial structure are crucial prerequisites for generating realistic 3D head avatars. In recent years, several 3D facial reconstruction methods, based on the seminal 3D Morphable Model (3DMM) \cite{Blanz2023morphable,Lou2021}, have shown impressive performance in both geometric and textural reconstruction with facial details such as expressions and wrinkles \cite{Feng2021learning, Danvevcek2022emoca}. However, they usually do not consider the details of non-facial and background regions, which are indispensable for high-fidelity head avatar generation, making general 3D facial reconstruction methods unsuitable for the head avatar Reenact task.

To deal with the above issues, we propose a real-time head avatar construction method named SelfieAvatar using just one short selfie video for training in this study based on the StyleGAN architecture, which can recover fine-detail texture of the face and completing no-face region and other elements near the face and background. Specifically, we construct a head avatar generation framework including StyleGan-based networks for face and non-face image generation respectively. Furthermore, we propose a mixed loss detail texture reconstruction model for head foreground reconstruction and avatar image generation during the adversarial generation training focusing on recovering high-frequency information and reconstructing high-dimensional perception features. In particular, we introduce a facial foreground region details loss for fine detail representation recovery, leveraging an Implicit Diversified Markov Random Field(ID-MRD) to constrain the detail differences between face foreground and ground truth for high-frequency information recovery. In addition, we compute the cosine similarity between the extracted multi-scale features of the generated image and the real image to generate a more realistic head avatar. Experimental and visualization results show the superiority of our method in real-time generating detailed head avatars from a monocular video over existing methods.

\section{Related Works}

\subsection{3D Face Reconstruction with 3DMM}
The 3D Morphable Model (3DMM) \cite{Blanz2023morphable}, is a pioneering method utilizing PCA\cite{abdi2010pca} to model facial appearance and geometry within a low-dimensional linear subspace, and has become the cornerstone of 3D face reconstruction in recent years\cite{Yu2014regression}. Early fitting-based methods aimed to estimate the parameters of a statistical face model and establish a 3D face appearance model to fit 2D face images through optimization and learning \cite{Blanz2003face}. Other regression methods tried to estimate the 3DMM parameters by solving a nonlinear optimization problem, establishing a correspondence between the points in a single image and the established 3D face model \cite{faggian2006active}. These methods generally tend to produce overly-smooth results due to the lack of details and variations. Numerous 3DMM variants extended several auxiliary models, such as lighting models, pose models, and albedo models, to capture more facial variation and details\cite{Yu2012facial}. More recently, some methods integrated the emotion recognition model into face reconstruction to enrich facial expressions \cite{Danvevcek2022emoca}. 
These 3D facial reconstruction methods from a monocular image typically ignore the control of non-facial areas such as the neck and shoulders, and it is difficult to separate the foreground and background. This limitation negatively impacts the realism of the generated head avatar.

\subsection{Head Avatar Creation from a Monocular Video}
Creating a head avatar from a monocular video, distinct from face replacement or performance capture and animation, involves generating photo-realistic portrait images of a target person, including their face and non-face area (hair, neck, shoulder, and background), using a video of another person as a reference. A widely adopted approach for avatar reconstruction involves Dynamic Neural Radiance Fields(NeRF) \cite{Gafni2021NeRF}, which combines volumetric rendering with a scene representation network based on a low-dimensional morphable model. NeRFBlendshape \cite{gao2022reconstructing} achieved facial reenactment by adopting several multi-level voxel fields as bases with expression coefficients and combining multi-level voxel fields. Other popular methods leveraged the power of GANs \cite{abdal2019image2stylegan} to produce high-fidelity head avatars combining latent space and face parameters. 
StyleAvatar \cite{wang2023styleavatar} enabled real-time avatar generation by utilizing multiple StyleGAN-based networks and implementing a compositional representation along with a sliding window augmentation method, but ignoring fine facial texture and high-dimensional perceptual features.

\section{Method}

\subsection{Overview}
\begin{figure*}[!ht]
    \vspace{-1.0em}
  \centering
  \includegraphics[width=0.85\linewidth]{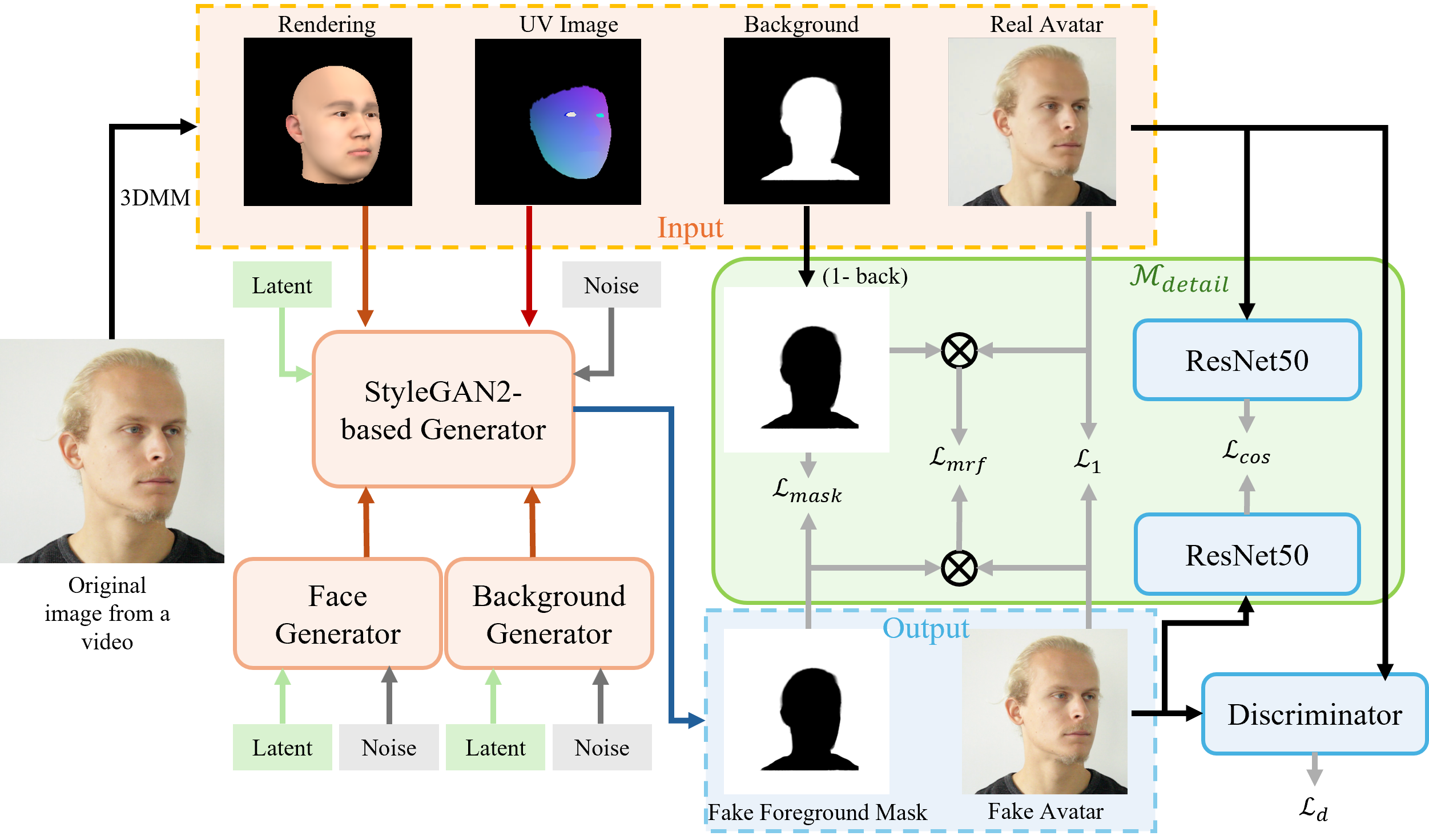}
  \caption{Overview of the detailed head avatar reconstruction method.}
  \label{fig_overview}
    \vspace{-1.0em}
\end{figure*}
Given a selfie video, we first carry out 3D face tracking \cite{wang2022faceverse} based on 3DMM to obtain the real face image and corresponding 3D face rendering image, UV image, and background image. To generate high-fidelity realistic head avatars with detailed faces and reasonable non-facial regions, we first perform two generation tasks in parallel for face and non-facial regions based on the StyleGAN generators \cite{karras2020analyzing} taking the latent vector and noise as inputs.  Then combining the generated face and background, we take them and the original rendering images and UV images as the input of the StyleGAN-based generator to generate a head avatar and corresponding head foreground mask. To recover the refined head avatar and avoid confusion between the background and the head foreground, we combine several training losses to construct a training strategy focusing on the head foreground image recovery and the whole avatar image generation. The overview pipeline is shown in Fig. \ref{fig_overview}. During the inference stage, through input of the face rendering image and UV map, we obtain the head avatar in real-time by the StyleGAN-based generator.

\subsection{Fine Head Avatar Generation}
We first adopt two StyleGAN generators $G_{fa}$ and $G_{b}$ for face region and background generation respectively. The face region $F_{fa}$ and background $F_{b}$ can be obtained by 
\begin{equation}
F_{fa} = G_{f}(z_{f}, \varepsilon _{f}),
\label{eq_1}
\end{equation}
and 
\begin{equation}
F_{b} = G_{b}(z_{b}, \varepsilon _{b}),
\label{eq_2}
\end{equation}
where $z_{f}$ and $z_{b}$ are latent vectors and  $\varepsilon _{f}$ and $\varepsilon _{b}$ means added noise. The final head image and corresponding foreground head mask are generated to recover the detailed texture of the head. We adopt a multi-scale StyleGAN-based generator from \cite{wang2023styleavatar} to perform generating, taking $F_{fa}$,  $F_{b}$ from Eq. \ref{eq_1} and Eq. \ref{eq_2}, original rendering image $I_{R}$ and UV image $I_{UV}$ as input: 

\begin{equation}
[I_{FM}, I_{A}] = G_{style}(F_{fa}, F_{b}, I_{R}, I_{UV}, z_{g}, \varepsilon _{g} )),
\label{eq_3}
\end{equation}
where $I_{A}$ is generated fake avatar image and $I_{fm}$ is corresponding foreground head mask. $z_{g}$ and $ \varepsilon _{g} $ are additional embedded latent vector and noise respectively.

\subsection{Detail Texture Reconstruction}
To create a finely detailed avatar foreground including non-face regions such as neck and shoulder for a realistic high-fidelity photo, we propose a detail reconstruction model $\mathcal{M}_{detdail}$ containing several additional losses for the generated avatar and foreground during the training, focusing on the foreground detailed texture reconstruction. 

\begin{equation}
\begin{array}{l}   
\mathcal{M}_{detdail}((I_{FM}, I_{back}), (I_{A}, I_{RA})) \\=\mathcal{L}_{mask}+\mathcal{L}_{mrf}+\mathcal{L}_{1}+\mathcal{L}_{cos}
\end{array} 
\end{equation}\label{eq_loss}
where $\mathcal{M}_{detdail}$ involves 4 parts losses for different fields about generated foreground $I_{FM}$, avatar image $I_{A}$ and corresponding groundtruth $I_{back}$ and $I_{RA}$. Specifically, we first minimize the difference between the fake foreground mask and groundtruth by

\begin{equation}
\mathcal{L}_{mask} = MAE(I_{FM}, (1-I_{back})),
\end{equation}\label{eq_4}
where MAE indicates the mean absolute error. $I_{back}$ is original real background mask. However, unlike the simple foreground mask, the head foreground in the image is full of detail and texture. Therefore, the above MAE and common L1 loss are not suitable for recovering foreground details. To train the detail reconstruction of the foreground, we also minimize
\begin{equation}
\mathcal{L}_{mrf} = MRF(I_{FM} \odot I_{A}, (1-I_{back}) \odot I_{RA} ),
\end{equation}\label{eq_5}
where MRF represents the Implicit Diversified Markov Random Field (ID-MRF) loss \cite{wang2018image}, which calculates the difference between the extracted features from different layers of VGG19.  In MRF, $\odot$ represents Hadamard product. We utilize ID-MRF loss to reconstruct high-frequency details of the avatar foreground. 

For the generated avatar images, we adopt the $\mathcal{L}_1$ loss to supervise the training. In addition, inspired by VGGFace2 \cite{cao2018vggface2}, we extract the high-dimensional perception features of fake avatar images and corresponding groundtruth by pre-training ResNet50\cite{he2016resnet} and calculate the cosine similarity between them as a training loss.

To sum up,  we minimize the total loss function including generator loss $\mathcal{L}_{g} $ and discriminator loss $\mathcal{L}_{d} $:

\begin{equation}
\begin{aligned}
\mathcal{L} &= \lambda_{mask}\mathcal{L}_{mask} + \lambda_{mrf}\mathcal{L}_{mrf} \\ & + \lambda_{1}\mathcal{L}_{1} + \lambda_{cos}\mathcal{L}_{cos} + \lambda_{d}\mathcal{L}_{d}  + \lambda_{g}\mathcal{L}_{g} 
\end{aligned}\label{eq_6a}
\end{equation}
where the $\lambda$ represents the weight coefficient of each item.

\begin{figure*}[!t]
\vspace{-1.0em}
  \centering
  \includegraphics[width=0.85\linewidth]{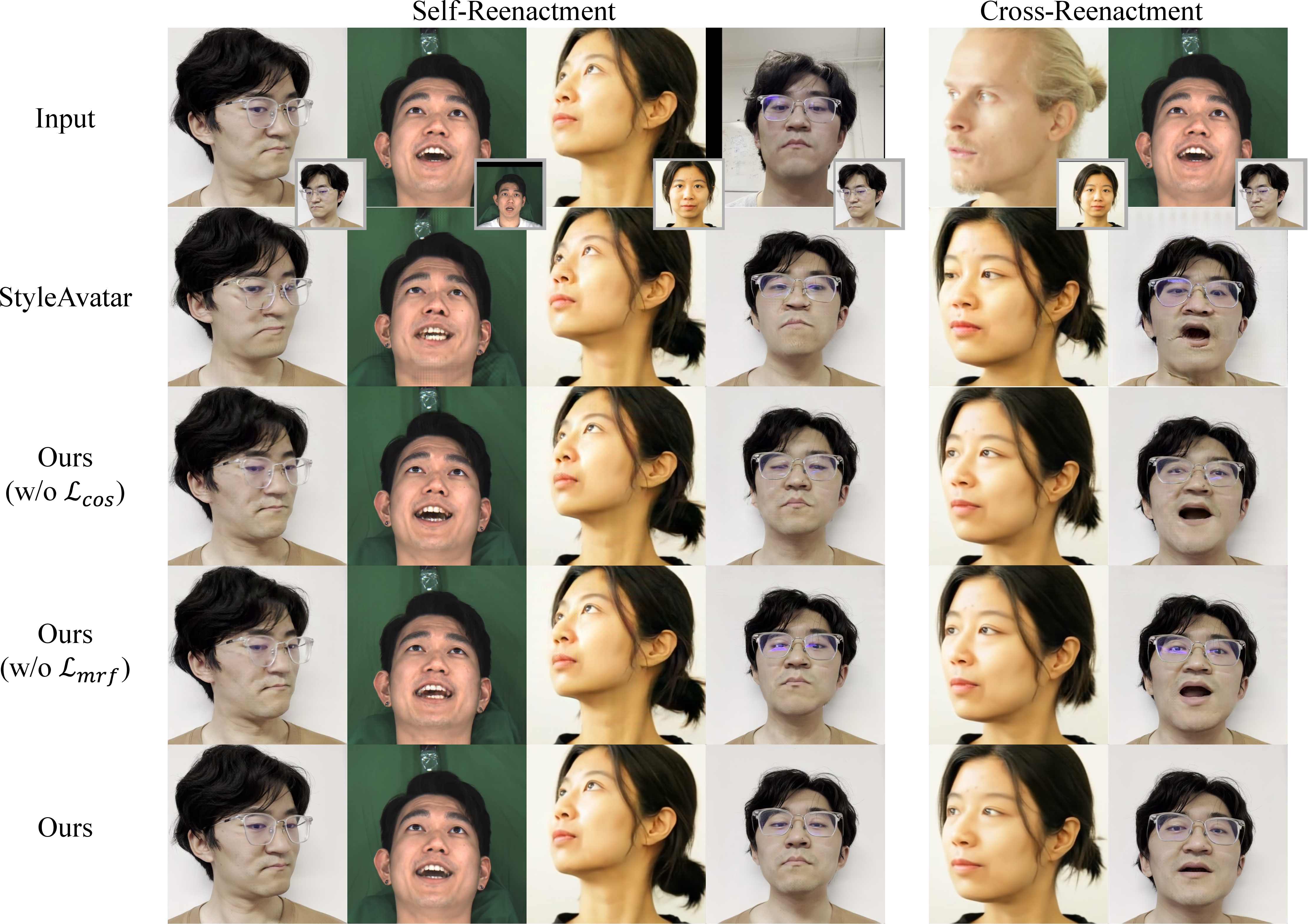}
  \caption{ Qualitative comparison results on head avatar self-reenactment and cross-reenactment.}
  \label{fig_result}
\end{figure*}

\section{Experiments}
\subsection{Implementation Details}
We perform our experiments including corresponding ablation studies on two types of tasks: self-reenactment and cross-reenactment. A about 3-minute selfie video of one identity is needed as training data. During testing, a different selfie video of the same identity is used for self-reenactment, while a video of a different identity is used for cross-reenactment. In addition, we furthermore perform the experiments on two face datasets from MEAD\cite{wang2020mead} and IMAvatar\cite{zheng2022avatar} to verify the effectiveness of the proposed method. In experiments, we empirically set $\lambda_{mask} = 3$, $\lambda_{mrf} = 5\times10^{-2}$, $\lambda_{1} = 1$, $\lambda_{cos} = 1$, $\lambda_{d} = 1$, and $\lambda_{g} = 1$ in Eq. \ref{eq_6a}. 

\subsection{Results}

\begin{table}[!t]
\caption{Quantitative evaluation results compared with head avatar reconstruction methods and two ablation studies.}
\centering
\label{tab_1}
\resizebox{0.47\textwidth}{!}{
\begin{tabular}{@{}c|cccc@{}}
\toprule
Methods      & SSIM$\uparrow$ & PSNR$\uparrow$ & LPIPS$\downarrow$ & FID$\downarrow$ \\ \midrule
IMAvatar     &  0.78    &  19.0    &   0.069    &   59.7  \\
StyleAvatar  &  0.82    &  23.4    &  0.055    &  27.4   \\
Ours (w/o $\mathcal{L}_{cos}$) & 0.86 & 26.4 & 0.031 &  16.6   \\
Ours (w/o $\mathcal{L}_{mrf}$) &  0.79    &  21.0   &   0.061   &  26.7   \\
Ours         &  $\mathbf{0.88}$    &  $\mathbf{28.5}$    &   $\mathbf{0.027}$    & $ \mathbf{13.7}$   \\ \bottomrule
\end{tabular}}
  \vspace{-2.0em}

\end{table}

\begin{figure}[!t]
  \centering
  \includegraphics[width=\linewidth]{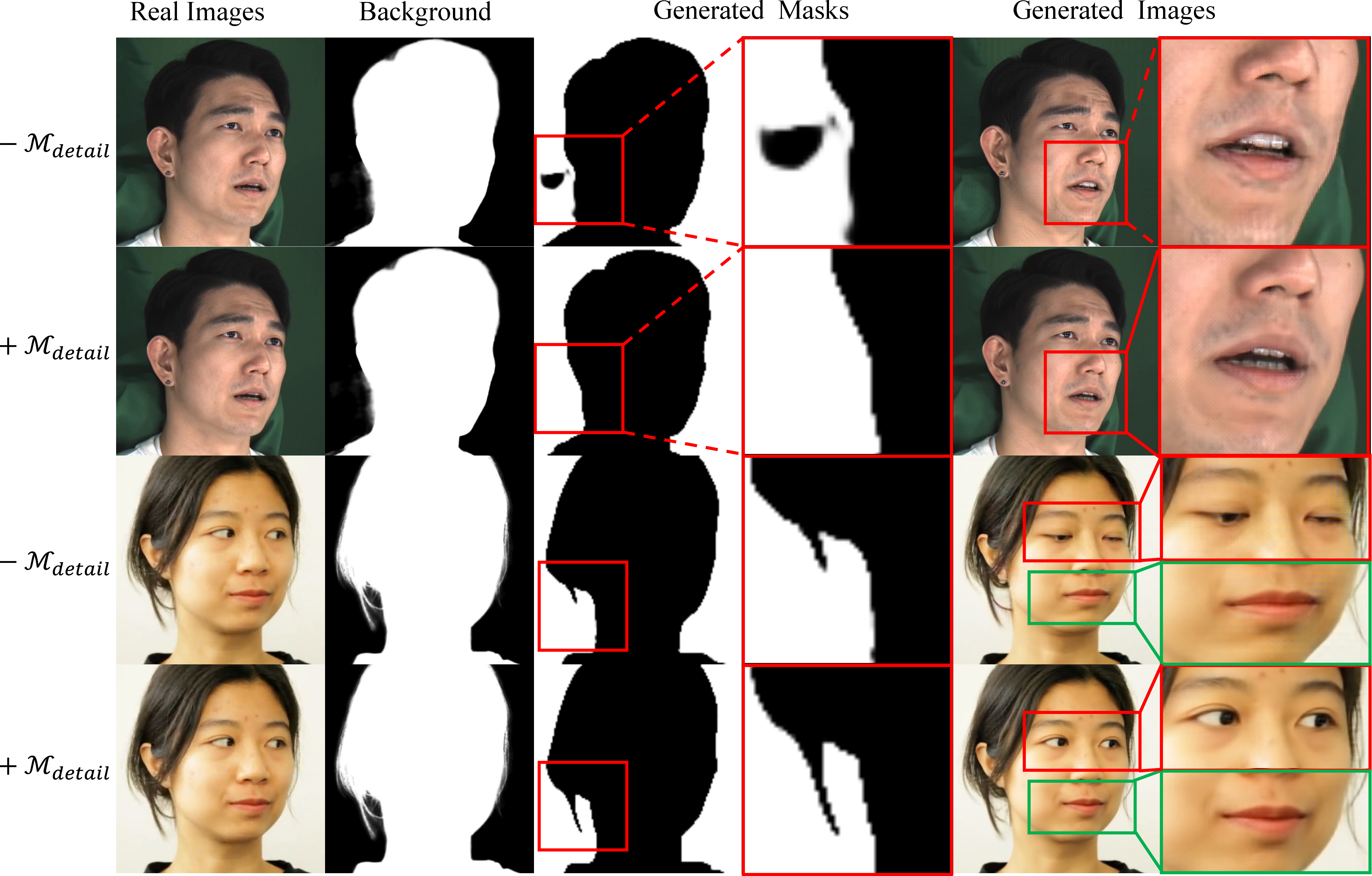}
  \caption{Visualization results for $\mathcal{M}_{detdail}$. From left to right:(1) ground-truth image of the head avatar, (2) corresponding background, (3) generated head masks with/without $\mathcal{M}_{detdail}$, and (4) output head avatar images. }
  \label{fig_detail}
  \vspace{-2.0em}
\end{figure}

In quantitative experiments, we compare our method with two state-of-the-art methods for head avatar self-reenactment from a selfie video. Each method is trained on the same selfie video and tested on another selfie video that is different from the training set but keeps the identity unchanged.  We quantitatively evaluate the Structural Similarity Index Measure(SSIM), Peak Signal-to-Noise Ratio (PSNR), and Learned Perceptual Image Patch Similarity(LPIPS) \cite{zhang2018lpips} and Fr$\rm \acute{e}$chet Inception Distance(FID) \cite{heusel2017fid} and provided the results in the Table.\ref{tab_1}.  
LPIPS computes the perceptual similarity between two input images by a backbone(AlexNet in this study).
It can be seen clearly that our method yields the best scores in terms of four metrics. The quantitative results of ablation studies for the Detail Texture Reconstruction model involving $\mathcal{L}_{cos}$ and $\mathcal{L}_{mrf}$ also verify the effectiveness of each component.

In qualitative experiments, we present visual comparisons of several samples for self-reenactment and cross-reenactment in Fig. \ref{fig_result} where in the first row of input images, the first and fourth columns are images from selfie videos, the second and sixth columns showcase images from the MEAD dataset, and the third and fifth columns are images from IMAvatar.
In self-reenactment, our method can reconstruct high-frequency head avatar images with fully detailed textures that are minimally different from the groundtruth.  In cross-reenactment, we obtain the most natural and realistic images with rich details and accurate posts. The visual results of ablations show that our detail texture reconstruction model can boost the detail generation accuracy such as hairs. 

To further demonstrate the efficiency of the proposed head detail reconstruction model, the visualization experimental results with/without $\mathcal{M}_{detdail}$ are presented in Fig. \ref{fig_detail}. From generated head masks in the third and fourth columns, more accurate head masks can be reconstructed with $\mathcal{M}_{detdail}$, avoiding inaccurate and rough edges. The last two columns show the finally generated head avatar images. This indicates that without $\mathcal{M}_{detdail}$, the details of the face such as eyes, teeth, and mouth are not well reconstructed. 

\section{Conclusion}
In this paper, we presented a fine-detailed head avatar reenactment method from a selfie video based on the 3DMMs and SytleGAN-based generator. We constructed a head avatar generation framework containing a detailed reconstruction model with several losses on different dimensions and fields for the head foreground texture recovery and the whole avatar image generation during training. Experimental results showed the effectiveness of our method on both self-reenactment and cross-reenactment. Visualization results showed that the detail reconstruction model had obtained more accurate head masks and better head avatar image generation than the previous methods, exactly in edge and texture details such as hairs and wrinkles.

%




\section{ETHICAL IMPACT STATEMENT} 
Participants in our study were informed about the research and data use, and we obtained 
informed consent before proceeding. Anonymized participant data were securely stored to 
protect privacy. Images of participants presented in this article received consent. The rest of 
the data comes from public datasets MEAD and IMAvatar for compared experiments. Our 
findings contribute to understanding social bonding and communication and may have 
applications in marketing and interpersonal skills training. We advocate for the ethical 
application of our research. We acknowledge the potential misuse of head avatar reenactment 
research for surveillance or manipulation. Ethical considerations and transparency are crucial.

{\small
\bibliographystyle{ieee}
\bibliography{egbib}
}

\end{document}